%% file: main.tex
\documentclass[runningheads, orivec]{llncs}
\usepackage[T1]{fontenc}
\usepackage{graphicx}
\usepackage{amssymb}
\usepackage{amsmath}
\usepackage{graphicx} 
\usepackage{tikz} 
\usepackage{listings}
\usepackage{caption}
\usepackage{tabularx} % for tables
\usepackage{booktabs}
\usepackage{multirow}
\usepackage{array}
\usepackage{hyperref} % links
\usepackage{orcidlink}
\usepackage{makecell}

\begin{document}
\usetikzlibrary {automata,shapes,arrows,positioning,arrows.meta, positioning, fit, shapes.geometric, calc, backgrounds}
\title{EBuddy: a workflow orchestrator for industrial human--machine collaboration}
\titlerunning{EBuddy: a workflow orchestrator for industrial human--machine collaboration}
\author{
    Michele Banfi\inst{1}\orcidlink{0009-0001-3594-7873} \and
    Rocco Felici\inst{1}\orcidlink{0009-0003-2157-3940} \and
    Stefano Baraldo\inst{1}\orcidlink{0000-0002-7941-8287} \and
    Oliver Avram\inst{1}\orcidlink{0000-0002-1228-5950} \and
    Anna Valente\inst{1}\orcidlink{0000-0003-3257-4482}
}
\authorrunning{M. Banfi et al.}
\institute{
    \href{https://www.supsi.ch/en/web/isteps/automation-robotics-and-machines}{Laboratory of Automation, Robotics and Machines (ARM)}, University of Applied Sciences of Southern Switzerland, Lugano-Viganello, Switzerland
    \email{\{michele.banfi, rocco.felici, stefano.baraldo, oliver.avram, anna.valente\}@supsi.ch}\\
}
 \maketitle              % typeset the header of the contribution
\begin{abstract}
%The abstract should briefly summarize the contents of the paper in 150--250 words.
This paper presents \emph{EBuddy}, a voice-guided workflow orchestrator for natural human–machine collaboration in industrial environments. EBuddy targets a recurrent bottleneck in tool-intensive workflows: expert know-how is effective but difficult to scale, and execution quality degrades when procedures are reconstructed ad hoc across operators and sessions. EBuddy operationalizes expert practice as a Finite State Machine (FSM) driven application that provides an interpretable decision frame at runtime (current state and admissible actions), so that spoken requests are interpreted within state-grounded constraints, while the system executes and monitors the corresponding tool interactions. Through modular workflow artifacts, EBuddy coordinates heterogeneous resources, including GUI-driven software and a collaborative robot, leveraging fully voice-based interaction through automatic speech recognition and intent understanding. An industrial pilot on impeller blade inspection and repair preparation for Directed Energy Deposition (DED), realized by human-robot collaboration, shows substantial reductions in end-to-end process duration across onboarding, 3D scanning and processing, and repair program generation, while preserving repeatability and low operator burden.

%This paper presents \emph{EBuddy}, a voice-guided workflow orchestrator for human--machine collaboration in industrial settings. The system addresses usability, knowledge transfer, and deskilling challenges in industrial repair workflows, here displayed by a use case of impeller blade inspection and repair via directed energy deposition (DED). Traditional workflows rely on expert operators for 3D scanning, data processing, and part program generation, resulting in low efficiency and high variability. EBuddy integrates domain-specific expert knowledge within a state machine–driven application designed to assist novice and intermediate operators in executing complex workflows, including autonomous collaborative robot (cobot) operations. The system enables fully voice-based interaction through automatic speech recognition and intent understanding. Experimental deployment in an industrial pilot cell demonstrates that EBuddy enables reliable know-how transfer, reduces cognitive load, and supports Industry~5.0 goals of human-centric, inclusive, and intelligent manufacturing.

\end{abstract}
%
%
%
%----------------------------------------------------------
\section{Introduction}
%----------------------------------------------------------
Industrial workcells increasingly combine heterogeneous digital tools (e.g., scanning and inspection software, data-processing pipelines, robot controllers) with collaborative automation~\cite{matheson2019}. In practice, performance is often constrained less by the robot’s physical capability than by the workflow decision space: knowing what step is valid next, choosing parameters, handling exceptions, and coordinating handovers between software and hardware~\cite{rodriguez2021}. These choices are usually clear to a small number of domain experts, yet they are hard to operationalize consistently across operators, shifts, and product variants.
This paper presents EBuddy, a voice-guided workflow orchestrator that represents procedures as Finite State Machine (FSM) recipes and uses the current state to frame and constrain operator commands. Interaction becomes a sequence of admissible decisions: the operator always has a stable view of the active state and available actions, and can progress through state-aware voice commands; workflows remain extensible via modular and nested FSMs without becoming unreadable. The approach is demonstrated in an industrial pilot on a turbomachinery impeller inspection and repair preparation pipeline, spanning GUI-driven scanning software, 3D data processing, and cobot-supported operations with quantitative and qualitative evidence indicating improved repeatability and reduced execution friction in a multi-tool setting.

\section{Related Works}
%----------------------------------------------------------

\subsection{Voice-based human-machine interaction}

Voice-based human–machine interaction has evolved from early keyword-based speech interfaces to context-aware, conversational systems, driven largely by deep learning advances in automatic speech recognition, language understanding, and speech synthesis~\cite{mehrish2023speechdl,guo2023speech}. Contemporary speech user interfaces now underpin applications ranging from mobile digital voice assistants and smart speakers to social robots.
Voice-based interaction is emerging as a key modality also in industrial settings, enabling operators to issue hands-free commands while maintaining situational awareness and manual dexterity on the shop floor. Recent work on collaborative robots shows that Automatic Speech Recognition (ASR) can achieve reliable speech recognition in noisy environments, even in the presence of technical vocabulary~\cite{avram2024advancingcom}. %Industrial speech interfaces treat voice as a direct command channel, whereas EBuddy uses workflow context to constrain interpretation to a closed set of valid actions.
%~\cite{seaborn2021voicehai,chen2023dva}.  REMOVED FROM ... social robots.

\subsection{Workflow orchestration in industry and HRC}
% \vspace{-\topsep}
%Recent years have witnessed a surge of research on integrating AI into industrial production, aiming to enhance flexibility, throughput, and quality in manufacturing systems. 
In recent years, foundation models demonstrated remarkable generalization capacities across tasks, and the recent introduction of standard protocols for agent-agent and agent-tool interaction~\cite{web:mcp} opened several new opportunities for integration with heterogeneous workflows. Yet, their direct deployment in manufacturing and industrial robotics remains constrained by a lack of contextual awareness, insufficient interpretability, and rigid integration with proprietary industrial software ecosystems. Consequently, researchers have increasingly focused on hybrid approaches that combine AI reasoning with symbolic, procedural, or knowledge-based control to ensure predictable behavior and compliance with safety and quality standards~\cite{brahman2024plasmamakingsmalllanguage,DBLP:journals/corr/abs-2109-14718,leanza2025conceptbotenhancingrobotsautonomy}.
Systems such as AI-enhanced CAD-CAM environments, context-aware programming assistants, and conversational interfaces for robotic programming aim to lower the cognitive and technical barriers to advanced production workflows~\cite{web:cadai,mallis2025cad,web:claude}. %These systems often embed expert knowledge through ontologies, digital twins, or reinforcement learning agents that encode task-specific best practices.
Still, many human-robot collaboration (HRC) systems assume skilled users, thus raising a barrier to broader adoption. %and do not explicitly address the deskilling of complex industrial workflows. Moreover, most current solutions are either tightly coupled to specific software stacks or limited to passive guidance and do not actively mediate between human intent and machine execution across heterogeneous tools.
Moreover, human–AI collaboration lies at the core of the emerging Industry 5.0 paradigm, as human agency and oversight are considered fundamental to guarantee the safety and transparency of human-robot collaborative systems~\cite{bonarini2020communication}. Within this domain, studies emphasize technologies that augment, rather than replace, human expertise~\cite{karpus2025human}. %Representative efforts include mixed-initiative systems that allow humans and machines to share decision authority, explainable AI frameworks that foster operator understanding and oversight, and adaptive interfaces capable of adjusting interaction modalities based on user skill level or task criticality.
%For example,~\cite{karpus2025human} investigates human cooperation with artificial agents, highlighting the role of perceived intent, mutual benefit, and trust.

These findings support the design of EBuddy as an explicit workflow orchestration partner, rather than a passive automation tool, that executes, monitors, and explains complex workflows while remaining accountable to the human operator.

%Recent advances in agentic systems, such as task-oriented digital assistants and agentic browsers, emphasize autonomous goal execution, context awareness, and user-in-the-loop control. EBuddy aligns more closely with this paradigm than with classical HRC, acting as an industrial workflow orchestrator that executes, monitors, and explains complex workflows while remaining accountable to the human operator.

%\subsection{Embodied and Multimodal Intelligence}

%Toyoda et al.~\cite{toyoda2021embodying} demonstrate how multimodal perception and embodiment enhance learning and task execution in human–robot systems. Similarly, Obayashi et al.~\cite{obayashi2025embodied} argue for embodied intelligence as a foundation for natural human–robot communication.

%----------------------------------------------------------
%\newpage
\section{Methodology}
% Describes what/how the system was built, objective details on features, architecture, implementation; no interpretation
%----------------------------------------------------------
This work addresses a fundamental challenge: users at different expertise levels often execute different workflows for the same industrial process, which creates both knowledge transfer gaps and process variability. Experts consistently reach high-quality outcomes by combining domain know-how with fluent tool use; intermediate users only partially master the problem and the toolchain and therefore deviate more often; beginners typically require sustained tutoring, tying up expert time that could be spent on higher-value work.

%users with different levels of domain expertise execute markedly different workflows for the same industrial process, creating knowledge transfer gaps and process variability. Expert operators reliably combine deep process knowledge with proficient tool use to reach high‑quality solutions, whereas intermediate users possess only partial understanding of both problem and tools and therefore deviate more frequently from optimal procedures. Beginners require extensive guidance and tutoring from more experienced colleagues, consuming substantial expert time that could otherwise be allocated to higher-value tasks. The \emph{EBuddy} solution presented in this paper is designed to support knowledge transfer between users — encompassing domain issues, software tools, and solution strategies — by formalizing optimal workflows defined by expert operators that can be consistently adopted by less experienced users.

EBuddy frames human–machine collaboration as a shared workflow orchestration framework, in which expert knowledge is explicitly encoded and operationalized through FSM-based workflows, multimodal interfaces, and distributed computation.
At runtime, the FSM provides an interpretable decision context, namely the current state and admissible actions, so operators can make sequential choices that remain consistent with process logic. By separating core orchestration from domain-specific artifacts (e.g., JavaScript Object Notation (JSON)-defined workflows, Python control scripts, instructional slides), the system enables asynchronous human–human knowledge transfer and progressive automation without removing expert oversight.

The overall system architecture, described in Fig.~\ref{fig:diagram}, is composed by the following components:
\begin{enumerate}
    \item Role-based interaction environment: manages user responsibilities (Sec.~\ref{sub:roles}).
    \item FSM architecture: implements different \textit{Workflows} (Sec.~\ref{sub:fsm}).
    \item \textit{Tool controls}: allow to control UI-based applications and physical tools such as robots (Sec.~\ref{sub:tools}).
    \item User interface (UI): informs users of process state, available commands, and context-sensitive guidance according to workflow progression (Sec.~\ref{sub:ui}).
    \item Language understanding pipeline, composed by ASR and context-aware intent recognition (Sec.~\ref{sub:lang}).
\end{enumerate}

% \begin{figure}[ht]
% \centering
% \includegraphics[width=1.0\linewidth]{Figures/overall-diagram.png}
% \caption{Collaboration between users and EBuddy to achieve optimal tool interactions, guided by domain expert knowledge encoded within EBuddy's workflow framework.}
% \label{fig:diagram}
% \end{figure}
\begin{figure}[ht]
\centering
\resizebox{\linewidth}{!}{\input{diagrams/arch}}
\caption{Collaboration between users and EBuddy to achieve optimal tool interactions, guided by domain expert knowledge encoded within EBuddy's workflow framework.}
\label{fig:diagram}
\end{figure}

% The EBuddy methodology addresses following key challenges in industrial human-machine collaboration:
% \begin{enumerate}

%     \item Skills deskilling through automation:
    
%     Fig.~\ref{fig:diagram} demonstrates how Fluently reduces expertise barriers by automating complex tasks that previously required specialized knowledge. The interactions needed for operating applications implementing 3D scanning, 3D data processing, and part program generation are mostly overtaken by EBuddy.

%     \item Device control:
    
%     The cobot control subsystem demonstrates EBuddy's device orchestration capabilities through algorithmic trajectory generation: real-time 3D measurements from a dual-camera system enable automatic computation of collision-free scanning paths (no programming needed).
    
%     \item Multimodal human-machine interface design:
    
%     Fig.~\ref{fig:diagram} shows how the shift from mouse-based to voice and intent-based interaction represents a natural user interface approach, making the system more intuitive and accessible.

%\end{enumerate}

%----------------------------------------------------------
\subsection{Knowledge transfer and user roles}\label{sub:roles}
%----------------------------------------------------------

%----------------------------------------------------------
%----------------------------------------------------------
EBuddy transfers application workflow knowledge from an expert—who defines workflows (FSMs with states and transitions), Tool controls, and contextual documentation—to technicians with different skill levels through a role-based interaction environment. Fig.~\ref{fig:diagram} demonstrates the different interactions with EBuddy of an advanced user and of a beginner/intermediate user. In this light,  EBuddy distinguishes three user profiles with progressively shifting responsibilities. Beginners rely on step-by-step guidance and mainly confirm the actions proposed by the default workflow: they observe EBuddy’s execution to learn core procedures, perform necessary physical interventions, and acknowledge safety-critical prompts. Intermediate users still use EBuddy as the primary interface but increasingly take local decisions, such as selecting among predefined options and tuning a limited set of parameters; they may enable automation for routine segments, handle common warnings, and monitor output quality to decide whether additional steps are needed. Advanced users act as process authors and maintainers: they formalize expert know-how by structuring the overall process into workflows and tool interaction primitives (widgets, scripts, checks, and rollback actions). They also establish defaults and safety guardrails, validate and refine execution strategies, and curate the instructional material that supports less experienced users. As proficiency increases, users move from observing and following guidance to adjusting parameters independently and eventually customizing or adapting the process, while EBuddy continues to execute routine operations reliably.

\subsection{Finite State Machine architecture}\label{sub:fsm}
%Expert users need a tool for performing workflow configuration.
As a workflow orchestrator for human-machine teams, EBuddy needs to feature:
%\vspace{-\topsep}
\begin{itemize}
    \item clear process state representation;
    \item downstream support to AI-based intent recognition and planning~\cite{10.1145/3729236};
    \item flexible and modular structure, allowing to easily define arbitrary task sequences involving humans, digital and physical agents;
    \item high human-readability, not only to support transparency, but also to directly serve as a guidance tool for the operator.
\end{itemize}
To satisfy these requirements, the FSM formalism was selected and implemented by the \emph{transitions}~\cite{web:transitions} Python library. Although behavior trees (BT) are widely adopted due to their modularity, reactivity, and hierarchical composition~\cite{avram2022generalized}, EBuddy prioritizes an explicit representation of the execution mode of the operation. In an FSM, the system’s status is captured by a single active state with a clear label, which can be directly exposed to the operator and logged for rapid diagnosis. This immediacy is particularly valuable in human--machine collaboration systems, where highly transparent operational states support supervision and safety assurance.  Moreover, the formalism of BTs can be difficult for beginners (the main target users of EBuddy) to grasp, as it requires learning \textit{depth-first} graph search and many low-level operators~\cite{coronado2020}.
%FSMs demonstrate superior interpretability and debuggability compared behavior trees (BTs)~\cite{avram2022generalized}, which require users to visually traverse complex hierarchical structures, while in FSMs they can immediately identify the active state and its label. This immediacy is particularly valuable in human--robot interaction systems, where unambiguous system state is critical for safety.
%The next figures allow to compare the spatial efficiency of BTs (Fig.~\ref{fig:gui-bt-coffee}) vs FSMs (Fig.~\ref{fig:gui-fsm-coffee}): note the reduced visual clutter and improved state transparency of the FSM approach compared to the BT.

%\begin{figure}[!htbp]
%\centering
%\includegraphics[width=1.0\linewidth]{Figures/gui-bt-coffee.png}
%\caption{A behavior tree illustrates a coffee-filling sequence. Adapted from Mohan~\cite{mohan2024developing}.}\label{fig:gui-bt-coffee}
%\end{figure}

%\begin{figure}[!htbp]
%\centering
%\includegraphics[width=1.0\linewidth]{Figures/gui-fsm-coffee.png}
%\caption{A finite state machine illustrates a coffee-filling sequence.}
%\label{fig:gui-fsm-coffee}
%\end{figure}
Finally, FSMs naturally accommodate constraint-based state reachability, allowing the Natural Language Understanding (NLU) (see Sec .~\ref{sub:lang}) module to restrict transitions to only valid, reachable states given the current system context. This alignment between natural language intent and valid state transitions creates an inherent enforcement mechanism for domain constraints without requiring external validation layers.

\subsubsection{Workflow definition}
\begin{figure}[ht]
\centering
\includegraphics[width=0.7\linewidth]{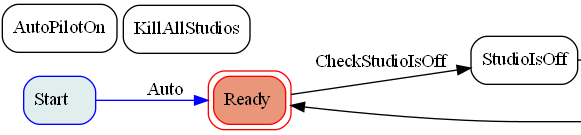}
\caption{The EBuddy preview tab enables users to initiate and configure the 3D Studio application in preview mode.}
\label{fig:gui-preview}
\end{figure}

Expert users can define new \textit{Workflows}, i.e., state machines dedicated to a specific task or goal. Workflows are defined by a JSON format, which defines states and transitions. 
%
%An expert user can leverage generative AI tools to automatically produce JSON workflow definitions for the Python transitions library from natural language descriptions of states and transitions. Fig.~\ref{fig:json} illustrates the resulting output: a JSON file that encodes the state machine by enumerating states and transitions, with each transition label mapped to a corresponding user voice command.FigureIsHere?
%
%\subsubsection{Autopilot}
Fig.~\ref{fig:gui-preview} shows a representation of a Workflow. In addition to Workflow-specific states and transitions, the expert may define \textit{jump states} (\emph{AutoPilotOn} and \emph{KillAllStudios} in the example), which are globally accessible from any state of the active state machine, and return control to the originating state upon completion. In particular, the \textit{autopilot} mode delegates the Workflow execution entirely to EBuddy, which is desirable in two antithetical scenarios: (1) a beginner in training lets the process run autonomously where possible, while learning from it; and (2) an expert user has optimized the Workflow, with custom states and transitions, to the point that it can be safely run without supervision.
%Once users have gained confidence through learning all workflow steps, they can activate the \emph{AutoPilotOn} command to delegate SM execution entirely to EBuddy. A common use case for autopilot is automating the preview tab, freeing the user to handle other operations. In autopilot mode, all steps execute autonomously, and EBuddy pauses if a warning or error occurs until the user resolves it. For example, the scanner may not be connected, so EBuddy will show the guide for connecting it. 

%The speed of the mouse can be easily changed by the user. If the user wants EBuddy actions to be quick, the set speed will be high. If the user wants to learn how to interact with the apps, the speed will be low, so that he can observe the mouse movements performed by EBuddy.

%\subsubsection{Transition execution flow}
%The actions that EBuddy supports during each transition are:
%\vspace{-\topsep}
%\begin{itemize}
%    \item \emph{TransitionDo}: actions executed when the transition is fired.
%    \item \emph{StateCheck}: conditions checked for evaluating the state validity.
%    \item \emph{Rollback}: actions executed when the transition is rolled back.
%    \item \emph{StateDo}: actions that are executed when the state is valid.
%\end{itemize}

EBuddy executes GUI interactions by controlling the mouse cursor; the cursor speed is user-configurable: a higher speed prioritizes fast execution, while a lower speed slows down motions to make actions easier to observe during learning. Actions may include Tool controls, API calls, or arbitrary Python function invocations. Moreover, FSMs support hierarchical composition: each FSM can invoke itself recursively or call other FSMs, enabling modular workflow organization.

% \begin{figure}[ht]
% \centering
% % \includegraphics[width=0.8\linewidth]{Figures/transition.png} % Old MB figure
% \input{diagrams/transition_conditions}
% \caption{Actions and conditions that can be mapped to a transition}
% \end{figure}

\subsection{Tool controls}\label{sub:tools}
EBuddy can control two types of Tools: \textit{Devices} and \textit{GUI-based applications}.

Devices include physical tools such as robotic units, cameras, actuators, or any other kind of device that can be integrated with a ROS2 interface. Tool control transitions trigger dedicated ROS2 nodes, which in turn manage the external hardware.

GUI-based applications can be controlled by EBuddy through programmatic interactions with graphical interface elements. %In the use case presented in Sec.~\ref{sec:case}, both the 3D scanner application and the part program generation software are orchestrated through this widget-based automation approach.
EBuddy executes predefined sequences of GUI operations — such as button clicks, form completions, and menu selections — to automate third-party application workflows that lack direct API access. To configure such interactions, an expert user captures a screenshot of the target UI element (e.g., a button in the window of the target application), saves it as a PNG file in the state machine directory, and references it as a transition action within the JSON workflow definition. When a transition associated with a GUI-based Tool is fired, EBuddy switches to the app and operates on the screen (e.g., clicking on the record button). Moreover, EBuddy can check the state of applications (e.g., searching the screen for a specific string) to update the state machine based on the app feedback.
 %Cobot scanning trajectories are computed algorithmically by exploiting real time 3D measurements of the impeller geometry acquired through an external dual‑camera system. The algorithm integrates spatial data of the impeller geometry, the fixture/support configuration, and the worktable constraints to automatically generate collision-free motion paths, eliminating manual trajectory programming and reducing setup errors. EBuddy assumes control of this process through the part program generation app, thereby simplifying tool interactions while preserving user discretion over key parameters.

%To support proficiency development, users can gain insight into the optimal scanning process by observing how the EBuddy algorithm orients the scanner relative to the impeller. This physical demonstration is reinforced by monitoring the standoff distance between scanner and impeller visualized in the 3D Studio app.

%The best way to operate scanning is to be able to highly overlap each acquired frame with the next one, without jumping (inadvertently jumping happens often in manual mode). This ``continuous'' behaviour is exactly what the algorithm does, and the result is that the 3D studio software can more easily calculate the interconnection between frames (registration), and thus the ``tracking lost'' tedious error message can be avoided. This way to operate can be learned by the user.

\subsection{User Interface}\label{sub:ui}

%\subsubsection{Informations streams}

\begin{figure}[ht]
\centering
\includegraphics[width=\linewidth]{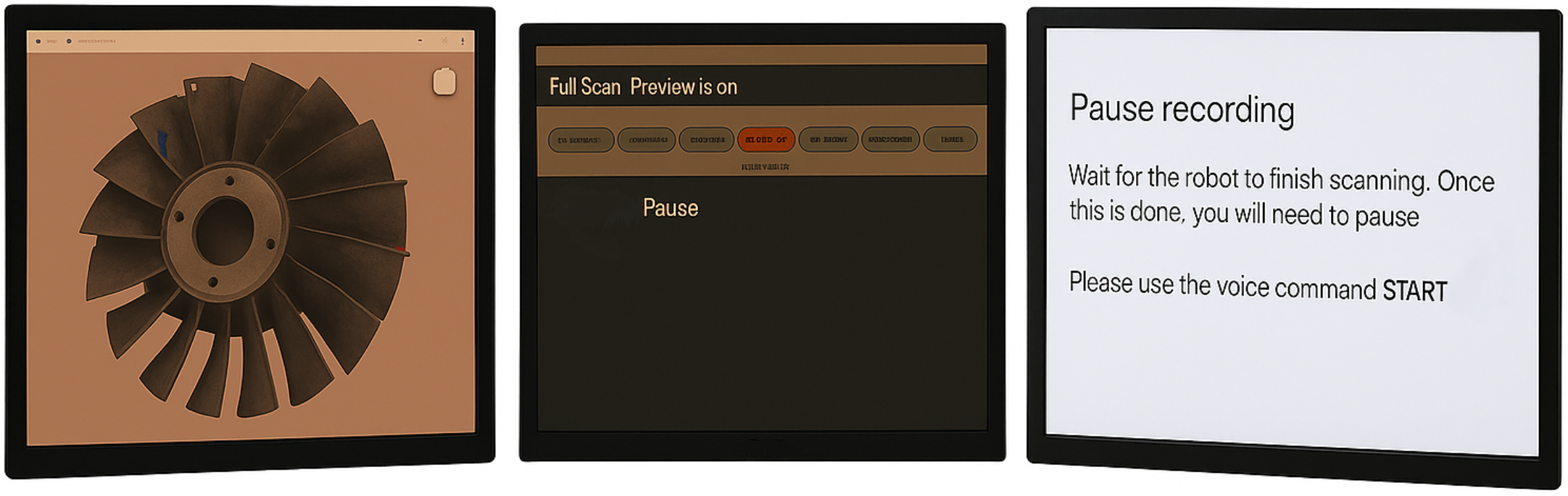}
\caption{Graphical information managed by EBuddy: GUI-based Tools (left), EBuddy GUI (center), Helper app (right).}
\label{fig:monitors}
\end{figure}

EBuddy exposes three types of graphical information (Fig.~\ref{fig:monitors}): %The multi-stream architecture enables selective attention allocation: depending on user type and workflow context, operators may focus intensively on one stream while disregarding others. The streams present the following information:FigureIsHere?

%\vspace{-\topsep}

\begin{enumerate}
    \item GUI-based tools: the applications controlled by EBuddy.%, which show specific results (a 3D scan, a processed mesh, a part program, etc).
    \item EBuddy GUI: a window displaying the overall workflow, including accessible transitions and states.
    \item Helper app: context-sensitive guidance slides.
\end{enumerate}

%\subsubsection{EBuddy GUI}

\begin{figure}[ht]
\centering
\includegraphics[width=1.0\linewidth]{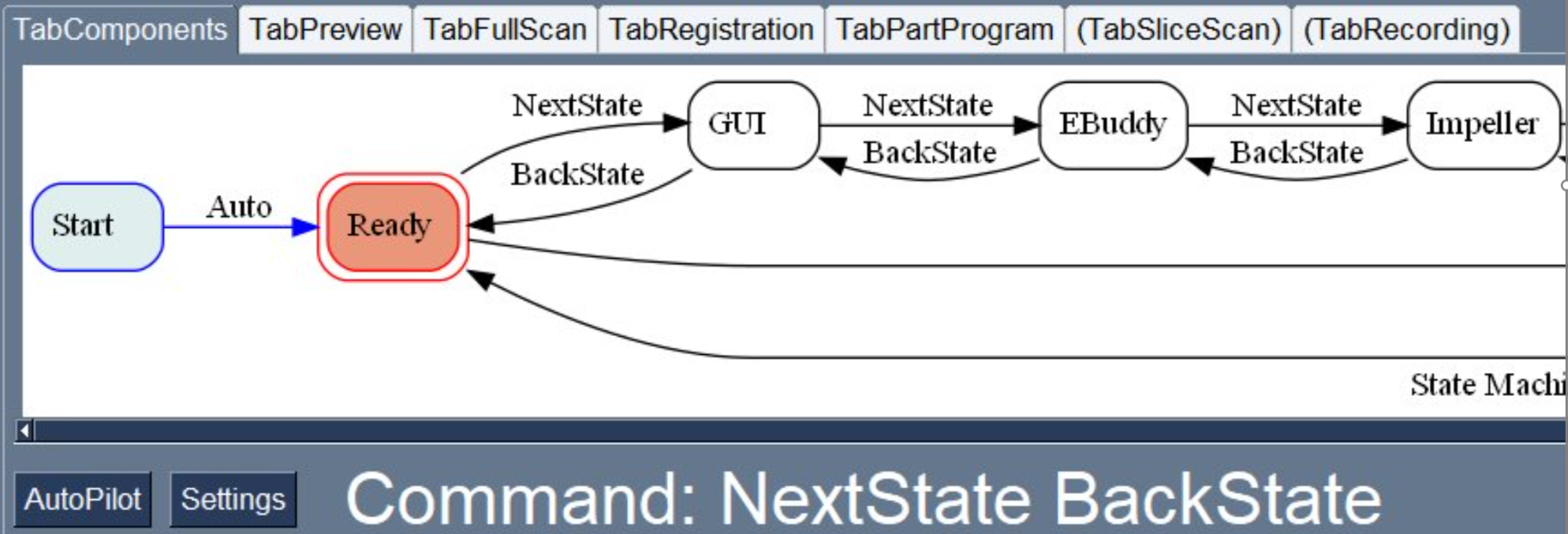}
\caption{EBuddy GUI is displaying the \emph{TabComponents} SM. The current state is \emph{Ready}, available commands are \emph{NextState} or \emph{BackState}.}
\label{fig:gui-tabs}
\end{figure}

EBuddy GUI displays all the available Workflows within tabs, arranging states in a left-to-right layout (Fig.~\ref{fig:gui-tabs}). States appear as rectangular boxes, with the current state highlighted in red. Voice commands are upon the transition arrows, enabling users to navigate through the workflow, loop back or jump out.

When the user pronounces a sentence that is classified as one of the available commands, EBuddy executes it, triggering either a GUI control or a Device control. In both cases, the user can observe the controller actions - screen interactions or robot movements - as planned by an expert, and learn when and how they are performed.

%Upon completing a "tab", the user can proceed to the next "tab" by navigating them from left to right; in this example, progression occurs from the ``TabComponents'' tab to the ``TabPreview'' tab. This left-to-right tab navigation enables completion of the full process, culminating at the final tab.

%\subsubsection{State visualization}\label{state}

%\begin{figure}[!htbp]
%\centering
%\includegraphics[width=0.6\linewidth]{Figures/gui-fsm.png}
%\caption{The \emph{Components} SM illustrates previous, current, and next states.}\label{fig:state}
%\end{figure}

%Fig.~\ref{fig:gui-tabs} shows a typical EBuddy state visualization for the help documentation SM. The current state is \emph{Ready} (highlighted in red), with the previous state \emph{Start} (blue, the initial startup state). From \emph{Ready}, the user can issue the \emph{NextState} voice command to advance to the \emph{GUI} state, which displays slides presenting the GUI. From the \emph{GUI} state, the \emph{BackState} command returns to \emph{Ready}, presenting the corresponding \emph{Ready} default slide.

\vspace{-20pt}

\subsection{Language Understanding}\label{sub:lang}%@rocco
The main modality of interaction with EBuddy is natural language. The understanding and execution of language instructions provide a flexible means for the operator to control the system; however, the interpretation of utterances is challenging due to the high degree of variability of language. Moreover, voice requests must be mapped exclusively to valid commands, i.e., state transitions accessible from the current state. 

The operator is equipped with a RealWear Navigator 500~\cite{web:skillworx} headset that captures voice commands in real time and streams them to a personal device via WebRTC. The voice signal is consumed fully on-device, to provide a higher level of privacy preservation~\cite{bussolan2025personalized}. In particular, the signal is pre-filtered by a Voice Activity Detection (VAD) module to distinguish silence from speech. Then, candidate speech chunks are sent to the ASR module, which is based on a pretrained base size Whisper model (Whisper~\cite{radford2023robust}); the model is also provided with a text prompt, which contains context-related technical vocabulary, highly improving word detection rates~\cite{avram2024advancingcom}. The operator's queries, now turned into text, are then forwarded to a desktop workstation via a communication bridge between an MQTT broker and ROS2. The operator's communicative intent is interpreted on the workstation. %As depicted in Fig.~\ref{fig:diagram}
During the application setup phase, the Natural Language Understanding (NLU) ROS node receives as input the complete structures of the available Workflows, i.e., the set of all possible interactions $t\in T$ that the operator can have with the system (transitions) in each phase $c$ of the process (states). During production, the user can access a closed set of commands that allow transitions through the state machine. Moreover, state-independent commands are always available; these commands allow general-purpose interactions, such as navigating the graphical interface, enabling the \emph{autopilot} function, or invoking the documentation. 
\begin{figure}[ht]
\centering
\input{diagrams/diagram}
\caption{Inputs and output of the language-based interaction for FSM navigation.}
\label{fig:in_out_lu}
\end{figure}
As depicted in Fig.~\ref{fig:in_out_lu}, the NLU node allows the interpretation of the operator's voice command by associating it with the transition $t$ having the highest degree of similarity. This mapping is strongly conditioned by the state $c$ of the system at the time of the interaction; in this way, the possible transitions are reduced exclusively to the set $\{{t_c}^1,{t_c}^2,\dots,{t_c}^n\}$ of transitions available from the current state $c$, thereby reducing the complexity of the classification problem. 

The choice of the correct transition among the available is based on various text similarity measures combined. In particular, lexical similarity between texts is evaluated by alignment and set distances, while semantic similarity is computed based on embedding comparisons.

\vspace{-10pt}
\subsubsection{Alignment and set-based distances}
The distance between two strings $a, b$ can be measured by the Levenshtein edit distance $D_{\text{Lev}}(a,b)$, defined as the minimum number of editing operations or minimum number of characters edited to get $b$ from $a$. Editing operations comprise: substitution of a character with another, insertion of a new character, or deletion.

Considering two strings as two sets of characters $A=\{t_a^1,t_a^2,\dots,t_a^m\}$, $B=\{t_b^1,t_b^2,\dots,t_b^n\}$, they can be compared by the Jaccard similarity (and thus defining the correspondent distance) as the ratio between the size of their intersection and the size of their union, i.e.
\begin{equation*}
S_{\text{Jac}}(a,b) = \frac{|A \cap B|}{|A \cup B|}, 
\qquad
D_{\text{Jac}}(a,b) = 1-S_{\text{Jac}}(a,b).
\end{equation*}

% \begin{equation}
%     S_{\text{Jac}}(a,b) = \frac{|A \cap B|}{|A \cup B|}, 
% \end{equation}

% thus defining the distance

% \begin{equation}
% D_{\text{Jac}}(a,b) = 1-S_{\text{Jac}}(a,b).
% \end{equation}

%and as distance:
%\begin{equation}
%    D_{\text{Jac}}(a,b) = 1-S_{\text{Jac}}(a,b) = 1 - \frac{|A \cap B|}{|A \cup B|} = \frac{|A \cup B|-|A \cap B|}{|A \cup B|}
%\end{equation}

\subsubsection{Vector embeddings and similarity function}
Given a pretrained language embedding model $f_\theta$ with parameters $\theta$, a latent embedding, i.e., a dense vectorial representation, of a text $a$ is defined as $x=f_{\theta}(a)\in \mathbb{R}^d$. 
To evaluate semantically similar texts, given two queries $a,b$ and the corresponding embeddings $x,y$ L2-normalized as $\hat{x}=\frac{x}{\lVert x \rVert_2}$, we can use as similarity function the cosine similarity:
\begin{equation*}
S_{\text{cos}}(x,y) = \hat{x}^\top\hat{y},
\qquad
D_{\text{cos}}(x,y) = 1-S_{\text{cos}}(x,y).
\end{equation*}
% \begin{equation}
%     S_{\text{cos}}(x,y) = \hat{x}^\top\hat{y} 
% \end{equation}
% and as distance
% \begin{equation}
%     D_{\text{cos}}(x,y) = 1-S_{\text{cos}}(x,y).
% \end{equation}
In practice, for creating the embedding model $f_\theta$, a pretrained transformer-based model is selected~\cite{wang2020minilmdeepselfattentiondistillation}. The model is a distilled version of an uncased sentence embedding model~\cite{devlin2019bertpretrainingdeepbidirectional} with 12 layers, 384 hidden size and 33M parameters. %; precisely, it is used a 6-layer version of it), 2.7x time faster than the base.
%Such model is trained on a large variety of textual data using self-supervised contrastive learning. This training framework not only involves a massive training with a next-token generation objective but also comprises a fine-tuning with a contrastive objective that helps distinguishing texts from different semantic areas. Such embedding model encodes textual data in a numerical vector form which capture the semantic information of such sentence. In our setting the transcription of the operator's query is encoded and compared via cosine similarity against available transitions in the current phase of the process. 

\subsubsection{Joint similarity assessment}
The different distance and similarity metrics are calculated individually but used jointly to uniquely evaluate the optimal transition. Available transitions $T$ in state $c$ are ranked for each metric $S_{\text{cos}}, D_{\text{Lev}}, D_{\text{Jac}}$ and the transition with the highest similarity and lowest distances is evaluated against a set of thresholds to be tuned during deployment. We define the optimal transitions according to each metric as:
\begin{equation*}
o_{D_{\text{Lev}}} = \arg\min_{t \in T} D_{\text{Lev}}(t),
\qquad
o_{D_{\text{Jac}}} = \arg\min_{t \in T} D_{\text{Jac}}(t),
\qquad
o_{S_{\text{cos}}} = \arg\max_{t \in T} S_{\text{cos}}(t).
\end{equation*}
Thresholds are defined during the deployment of the cell to $\tau_{S_{\text{cos}}}=0.5$ for the semantic similarity and $\tau_{D_{\text{Lev}}}=2,\tau_{D_{\text{Jac}}}=0.3$ for lexical distances.
% During the deployment of the cell thresholds are defined to $\tau_{S_{\text{cos}}}=0.5$ for the semantic similarity and $\tau_{D_{\text{Lev}}}=2,\tau_{D_{\text{Jac}}}=0.3$ for lexical distances. % However, although the robot is collaborative, potential false positives occurred during the intent classification can result in unexpected movements of the robot. Such movements increase the risk of collisions, slow down the normal course of the scanning process, and could compromise the quality of the final result. For this reason, transitions that produce rapid or large movements of the robot are associated with more stringent thresholds, at the cost of limiting the ability to generalize to different language nuances: $\tau_{S_{\text{cos}}}=0.5,\tau_{D_{\text{Lev}}}=2,\tau_{D_{\text{Jac}}}=0.25$.
%
% Defining the thresholds $\tau_{D_{\text{Lev}}},\tau_{D_{\text{Jac}}}$ for lexical distances and $\tau_{S_{\text{cos}}}$ for the semantic similarity, 
The unique optimal transition that underlies the safety thresholds is defined as:
\begin{equation}
o =
\begin{cases}
o_{D_{\text{Lev}}}
& \text{if } D_{\text{Lev}}(t) \le \tau_{D_{\text{Lev}}} \\[6pt]

o_{D_{\text{Jac}}} 
& \text{if } D_{\text{Lev}}(t) > \tau_{D_{\text{Lev}}} \;\land\; D_{\text{Jac}}(t) \le \tau_{D_{\text{Jac}}} \\[6pt]

o_{S_{\text{cos}}} 
& \text{if } D_{\text{Lev}}(t) > \tau_{D_{\text{Lev}}} \;\land\; D_{\text{Jac}}(t) > \tau_{D_{\text{Jac}}} \\
& \quad \land\; o_{S_{\text{cos}}} = o_{D_{\text{Lev}}} \;\land\;  S_{\text{cos}}(t) > \tau_{S_{\text{cos}}} \\[6pt]

\mathrm{False}
& \text{otherwise}
\end{cases}
\end{equation}

%----------------------------------------------------------
\section{Case study: cobot-assisted 3D scanning and analysis for metal part repairing}\label{sec:case}
%----------------------------------------------------------

%----------------------------------------------------------
%\subsection{Use case introduction}
%----------------------------------------------------------

%In this section, we describe a pilot implementation of EBuddy for a scan-and-repair collaborative robotic workcell, as shown in Fig.~\ref{fig:pa_cell}.

%\subsubsection{Background}
Vision-based 3D geometry reconstruction (3D scanning) requires moving a scanning device, e.g., fringe projection or photogrammetry, along the part surface. The full workflow, from acquisition to mesh optimization, demands high expertise in both sensor handling and the associated software. Hand-guided scanning can cause physical strain due to prolonged mid-air manipulation of a heavy device, while robot-guided scanning requires repeated interactions with control software to ensure adequate surface coverage.

A key application is metal part repair: additive manufacturing processes such as Direct Energy Deposition (DED) can restore damaged high-value components, supporting economic and environmental goals. Accurate capture of the as-is geometry and its comparison to the target geometry is therefore essential. However, such components often involve complex workflows. For instance, impellers have multiple radially arranged blades that can be thin (under 10 mm) and separated by deep, narrow voids—features that are difficult to capture with standardized robot scanning motions.

\begin{figure}[ht]
\centering
\includegraphics[width=0.7\linewidth]{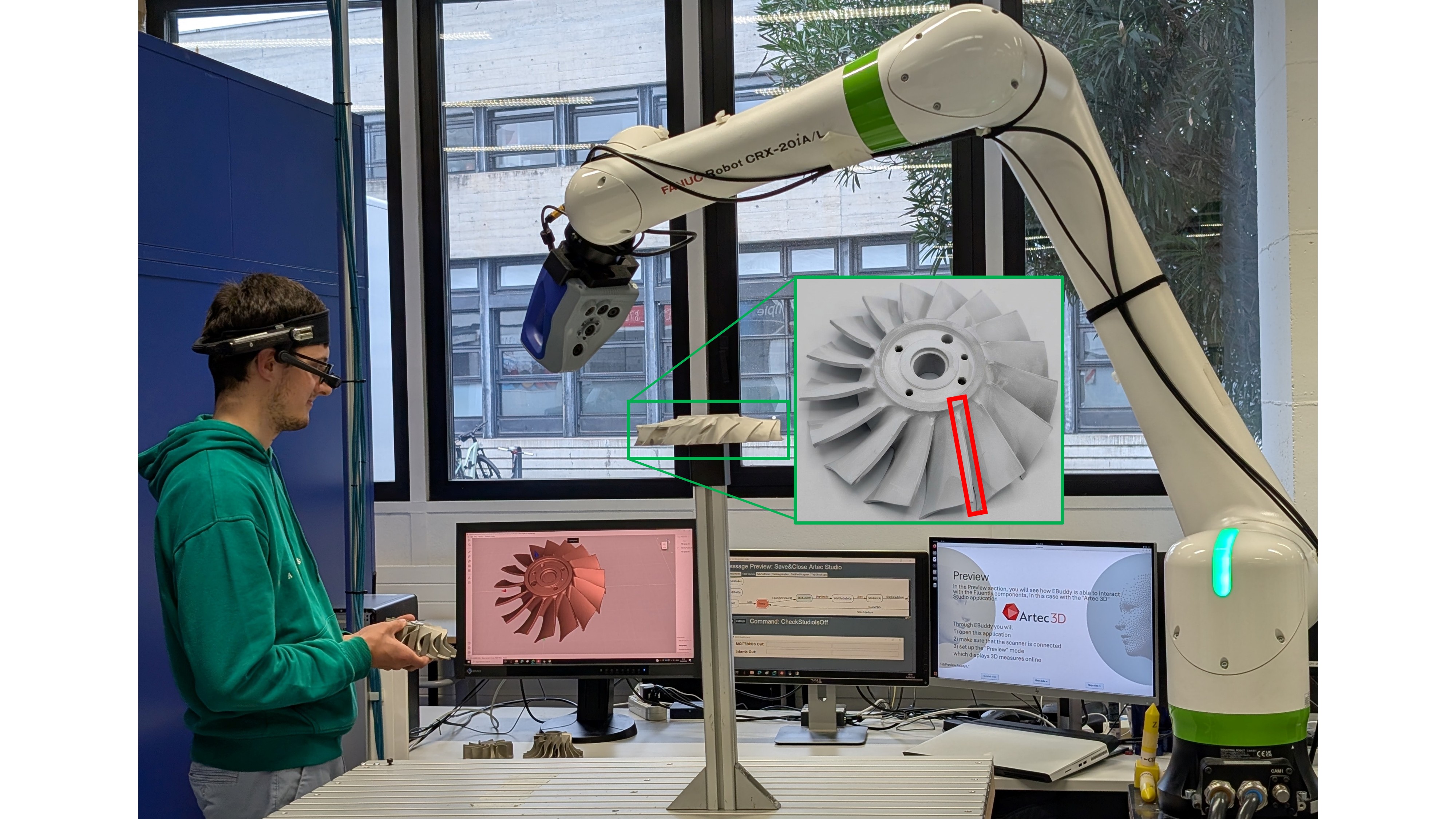}
\caption{Hardware configuration for impeller repair. Overview of the work-cell with an inset showing a detailed view of the impeller and the designated repair region.}
\label{fig:pa_cell}
\end{figure}

The following sections detail the EBuddy implementation of a cobot-assisted scan-and-repair process of an impeller for turbomachinery (Fig.~\ref{fig:pa_cell}).

\vspace{-30pt}

%----------------------------------------------------------
\subsection{Tools}\label{subsec:tool}
%----------------------------------------------------------

\subsubsection{Helper App}

The application functions as a voice-controlled slide viewer supporting the following commands: slide navigation (next/previous), confirmation (ok), skip mode (bypass detailed content), and detail mode (expand content). The helper app is driven by EBuddy's current state, enabling context-aware presentation: it displays slides whose names correspond to the active state name, ensuring workflow-specific guidance.

\subsubsection{3D Studio App}

The \emph{3D Studio App} refers to Artec Studio 16 software running on a Windows 11 desktop PC. It is employed for 3D data acquisition, processing, and scanner control of the Artec Spider handheld scanner — a device mounted on the cobot and designed for CAD and reverse-engineering applications. EBuddy executes GUI automation by manipulating interface elements and continuously monitoring visual feedback to identify processing completion.

\subsubsection{Cobot}

The system employs a Fanuc CRX-20i cobot integrated with ROS2 Humble for communication. The cobot is fixtured externally to the work table to optimize workspace utilization and reach efficiency, where an elevated fixturing for impellers enables the cobot to access the part from nearly all directions. %EBuddy transmits the motion commands to the robot controller via ROS2.

\subsubsection{Part Program Generation App}
This application supports the entire part program generation process. It displays the impeller CAD, the measured mesh, the difference between them, alignments, parameters input, final visualization and part program generation. %EBuddy interacts with this app using the same approach as with 3D Studio.

\subsection{Workflows}
\begin{table}[htbp]
\centering
\footnotesize
\setlength{\tabcolsep}{4pt}
\renewcommand{\arraystretch}{1.1}
\begin{tabularx}{\linewidth}{
|>{\raggedright\arraybackslash}p{2cm}|
>{\raggedright\arraybackslash}p{2.8cm}|
>{\raggedright\arraybackslash}p{2.8cm}|
>{\raggedright\arraybackslash}X|
}
    \hline
    \textbf{Workflow} & \textbf{Typical operator decisions} & \textbf{EBuddy \newline orchestration (examples)} & \textbf{Shared recovery strategies} \\
    \hline
    \textbf{Components 
    }& Choose guidance granularity; decide when to request help; confirm readiness to start
    & State-linked guidance via Helper App; structured  repeat / expand / skip actions
    & \textbf{Re-ground \& resume}: replay last relevant guidance snippet/state,then continue. \\
    \hline
    \textbf{Preview 
    }& Check prerequisites (scanner/software ready); choose autopilot vs step-by-step 
    & Launch/configure 3D Studio App; readiness checks; enforce required confirmations 
    & \textbf{Block--fix--retry}: stop on missing prerequisites, troubleshoot, retry transition.\\
    \hline
    \textbf{Full Scan 
    }& Select scan strategy/parameters; judge completeness; decide proceed vs refine
    & Guides acquisition order; constrain valid  actions; coordinates cobot viewpoint routine 
    & \textbf{Quality loop}: evaluate quality and $\rightarrow$ repeat a view or branch to refinement (Scan slicing) until criteria are met. \\
    \hline
    \textbf{3D Processing 
    }& Choose basic vs advanced path; accept vs rework based on mesh quality
    & Drives GUI sequence (merge/registration/fusion/export); keeps active step  and valid commands visible
    & \textbf{Local rollback}: re-run the minimal failed sub-step (e.g., re-register, re-fuse); escalate upstream to re-scan if needed. \\
    \hline
    \textbf{Scan Slicing 
    }& Select region of interest (ROI) to refine; decide sufficiency; iterate vs fallback
    & Bounded refinement sub-workflow; constrained  actions
    & \textbf{Iterative micro-loops}: repeat ROI scans; fallback to Full Scan if unresolved (e.g.,  gaps cannot be closed). \\
    \hline
    \textbf{Part Program Generator 
    }& Select parameters (e.g., region, layer thickness); validate output; regenerate vs proceed
    & Orchestrates external GUI pipeline (alignment/closure/deviation/clustering/export)
    & \textbf{Rollback and rerun}: adjust parameters and regenerate; route upstream on data issues.\\
    \hline
\end{tabularx}
\caption{Impeller repair Workflows.} 
\label{tab:workflow_structured}
\end{table}

In the scan-and-repair pilot, EBuddy operationalizes the end-to-end process as a set of modular FSM Workflows that coordinate the Tools introduced in the previous Sec.~\ref{subsec:tool} Tab.~\ref{tab:workflow_structured} briefly summarizes these Workflows by contrasting the typical decision points left to the operator with the corresponding EBuddy orchestration steps and shared recovery patterns. 

\section{Discussion and preliminary results}
% Reports what was found (data/metrics), Raw evidence, figures/tables; minimal explanation, "EBuddy achieved 95% task autonomy in tests."
%----------------------------------------------------------
In an initial evaluation campaign, six participants executed the scanning and repair preparation workflow in two conditions: Manual baseline (under continuous human expert guidance and training) and EBuddy-assisted (voice-guided orchestration with embedded expert workflow; human expert is required only for exceptional cases outside EBuddy's recovery logic). The manual baseline is workload-intensive for inexperienced users and inherently variable: operators must manually execute multiple tasks and coordinate several tools (scan planning and execution, tracking-loss recovery, scan-quality judgment, scan-to-CAD alignment, repair-volume definition, and DED program generation), incurring non-trivial cognitive and physical demands throughout the process. Furthermore, the baseline scenario is also expert-intensive, since novices typically require extensive up-front training and frequent synchronous assistance to complete the workflow correctly. 
EBuddy targets these bottlenecks by grounding spoken intent in the current process state and enforcing an FSM decision frame that exposes only admissible actions and recovery steps, thereby shifting execution from ad hoc procedure reconstruction to state-aware progression within a validated recipe. In this instantiation, the beginner mainly consumes and practices the encoded procedure (step-wise guidance and embedded checks), while the expert moves from continuous tutoring and process integration to authoring workflow artifacts (FSM recipes, scripts, and instructional content) and limited exception handling.
Operational impact was quantified with time measurements over the full job-preparation chain required to produce a repair-ready dataset and program. This spans three macro-phases: (i) Training and onboarding (scanner + software + workflow basics), (ii) Scanning (setup, acquisition, registration and processing, quality check and potential re-scan), and (iii) Repair-program generation (launching and completing part-program generation).
For each phase, operator hands-on time, synchronous expert assistance time, and system-executed time (state-machine modules performing/teaching steps) were separately recorded. 
% Proposta MB
%For each phase operator hands-on time, synchronous expert assistance time, expert EBuddy configuration time (for prepararing workflows, tool controls, cobot paths and slides) were separately recorded. 

% Version MB
As shown in Tab.~\ref{tab:workflow-efficiency}, user time is reduced from 868 min (manual) to 320 min (EBuddy), corresponding to a reduction of approximately 63\%. During manual operation, the expert accompanies users, so their work times are the same. With EBuddy, the expert invests 770 min in initial configuration but subsequently accompanies users for only 30 min total. As user count increases, expert time savings scale proportionally, freeing capacity for other tasks.
% Version OA
%As shown in Tab.~\ref{tab:workflow-efficiency}, EBuddy reduced hands-on user time from 145 min to 53 min (-64\%), and-critically for industrial scalability-reduced total time including expert effort from 867 min to 217 min (-75\%). The main driver was the collapse of synchronous expert involvement from 454 min to 57 min (-87\%), with most of the time concentrated around the assessment of the scan quality, part program generation and recovery actions.  In parallel, training time decreased from 289 min to 92 min (-68\%), consistent with the system’s ability to provide just-in-time, state-dependent guidance rather than prolonged tutoring. 

Collectively, the results show that EBuddy’s workflow orchestration and decision framing do not merely accelerate execution; they reallocate effort away from expert bottlenecks, improving repeatability and enabling higher throughput in settings where domain expert availability is the limiting resource.

% Version OA
% Manual vs EBuddy comparison table
%\begin{table}[ht]
%\centering
%\caption{Comparison of Manual and EBuddy workflow efficiency.}
%\label{tab:workflow_efficiency}
%\begin{tabular}{|p{6cm}|c|c|c|}
%\hline
%\textbf{Phase} & \textbf{Manual} & \textbf{EBuddy} & \\
%\textbf{(workflow macro-operations)} & \textbf{Total time (min)} & \textbf{Total time (min)} & %\boldmath$\Delta$ \\ \hline
%1. Training/onboarding phase & 289 & 92 & -68\% \\ \hline
%2. Scan and 3D processing phase & 518 & 101 & -80\% \\ \hline
%3. Repair program generation phase & 60 & 24 & -60\% \\ \hline
%Total time (incl. expert) & 867 & 217 & -75\% \\ \hline
%Operator hands-on time \newline (average across 3 users, all 3 phases) & 145 & 53 & -64\% \\ \hline
%Synchronous expert time \newline (sum across 3 users, all 3 phases) & 454 & 57 & -87\% \\ \hline
%EBuddy configuration time by expert  \newline \newline & - & 770 & - \\ \hline
%\end{tabular}
%\end{table}

% Version MB
% MB Manual vs EBuddy comparison table, in double columns
% Grezza stima del EBuddy configuration time, che in verità potrebbe essere ben piu' alto:
% (70 slides, 6 SM, 2 cobot paths)
% (70 slides * 5 min = 350 min + 6 SM * 60 min + 2 cobot paths * 30 min = 770 min)
\begin{table}[htbp]
\centering
%\begin{tabular}{|p{6cm}|c|c|c|c|c|c|}
\begin{tabular}{|p{4.5cm}|c|c|c|c|c|c|}
\hline
\textbf{Workflow Phase} & \multicolumn{2}{c|}{\textbf{Manual Op.}} & \multicolumn{2}{c|}{\textbf{EBuddy Op.}} & \boldmath$\Delta$ & \boldmath$\Delta$ \\
\textbf{} & \textbf{Users} & \textbf{Expert} & \textbf{Users} & \textbf{Expert} & \textbf{Users} & \textbf{Expert} \\ \hline
1. EBuddy configuration phase & 0 & 0 & 0 & 770 & -\% & -\% \\ \hline
2. Training/onboarding phase & 289 & 289 & 154 & 30 & -47\% & -90\% \\ \hline
3. Scan and 3D processing phase & 518 & 518 & 142 & 0 & -73\% & -100\% \\ \hline
4. Repair program generation phase & 60 & 60 & 24 & 0 & -70\% & -100\% \\ \hline
Total time (min) & 868 & 868 & 320 & 800 & -63\% & -8\% \\ \hline
\end{tabular}
\caption{Comparison of Manual and EBuddy workflow efficiency.}
\label{tab:workflow-efficiency}
\end{table}
\vspace{-30pt}
Preliminary observations indicate that the cell remains operable even by users with a limited technical background, suggesting low entry requirements and a fast learning curve. This points to \emph{skill democratization}: EBuddy externalizes critical decision points into an explicit, guided recipe, enabling beginners to reproduce strategies typically applied by advanced users. Participants reported lower perceived workload during scanning (partly due to fewer awkward postures), high flexibility/adaptability, and low programming effort. They also perceived shorter setup time for new parts and less time lost to 3D data reprocessing and re-scanning, consistent with decision-guided orchestration that enables higher throughput without increasing operator burden.
To keep decision-making tractable as recipes expand, EBuddy structures the workflow as nested state machines: each layer exposes only the decisions relevant at that granularity, while subordinate FSMs encapsulate detailed steps. In practice, the operator always sees the active state and the admissible commands, so the \emph{decision frame} remains stable even in large state spaces; when more context is needed, voice-controlled navigation allows quick traversal without losing contextual grounding. This modular composition is the main scalability lever beyond the specific repair case: new workflows, tasks, variants, and decision points can be introduced as callable modules within the same orchestration pattern, preserving a stable decision frame and operator usability as process scope grows. 

%----------------------------------------------------------
\section{Conclusions and Future Work}
EBuddy advances HRC from voice command to decision-centric workflow orchestration: speech input is interpreted within an explicit, state-grounded decision frame, and intent-driven actions are executed through accountable, modular FSM recipes that remain readable, verifiable, and extensible. In the industrial pilot, this paradigm translated expert practice into a repeatable interaction substrate, supporting rapid onboarding, skill democratization, and consistent execution across heterogeneous tools (GUI-based software and cobot operations) without requiring API-level integration. The result is a pragmatic step toward Industry 5.0 human-machine collaboration: operators retain agency over workflow decisions, while Ebuddy enforces orchestration with shared recovery strategies, ensuring traceable progression across complex procedures. 

Future work will: (i) involve a broader and diverse sample of participants, to clearly quantify the benefits of the approach; (ii) scale the validation beyond the current scenario through controlled studies with quantitative quality/error metrics and benchmarking against conventional GUIs and alternative orchestration formalisms; and (iii) strengthen robustness by adding verification modules (e.g., mesh-closure/coverage checks) and systematic exception handling. From a deployment perspective, a reusable library of workflow modules and authoring aids needs to be developed to reduce configuration overhead and accelerate transfer to new processes, while exploring adaptive user modeling to personalize guidance and autonomy levels.

\begin{credits}
\subsubsection{\ackname} This work was funded by Horizon Europe project \textit{Fluently} (grant ID 101058680).

\subsubsection{\discintname}
The authors have no competing interests to declare that are relevant to the content of this article.
\end{credits}
\bibliographystyle{splncs04}
\bibliography{bibliography}

\end{document}

%% file: diagrams/arch.tex
% \documentclass[tikz,border=8pt]{standalone}
% \usepackage{tikz}
% \usetikzlibrary{arrows.meta, positioning, fit, calc, backgrounds, positioning}

% \begin{document}
\begin{tikzpicture}[
    every node/.style={font=\sffamily},
    font=\sffamily,
    dbl/.style={draw, {Stealth[length=1.7mm]}-{Stealth[length=1.7mm]}},
    box/.style={rectangle, draw, rounded corners=2pt, minimum width=3cm, minimum height=1cm, text centered, font=\sffamily, align=center},
    peach/.style={draw, rectangle, rounded corners=2pt, fill=orange!15, align=center},
    state/.style={draw, circle, fill=orange!15, minimum size=0.62cm, inner sep=0pt, font=\sffamily\scriptsize},
    smalltext/.style={font=\sffamily\scriptsize, align=center},
    tinytext/.style={font=\sffamily\tiny, align=center},
]

% Palette and reusable dimensions
\def\peach{orange!15}

% People icons
\node[circle, draw, fill=black, minimum size=0.30cm, inner sep=0pt] (headL) at (-7.05,2.5) {};
\node[draw, fill=black, rounded corners=6pt, minimum width=0.55cm, minimum height=0.32cm, rotate=18] (bodyL) at (-7.05,2.2) {};
\node[tinytext, text width=1.12cm] (expertLabel) at (-7.05,1.2)
  {Domain\\experts\\define\\workflows,\\tool controls,\\cobot paths\\and slides};

\node[circle, draw, fill=black, minimum size=0.30cm, inner sep=0pt] (headR) at (6,0.98) {};
\node[draw, fill=black, rounded corners=6pt, minimum width=0.55cm, minimum height=0.32cm, rotate=-18] (bodyR) at (6,0.68) {};
\node[tinytext, text width=1.0cm] (userLabel) at (6,0)
  {Users\\adopt\\EBuddy\\workflows};

% Main banner
\node[peach, minimum width=7cm, minimum height=0.5cm] (role) at (0,2.65)
  {\small{(1) Role-based interaction environment}};

% digraph
\node[state, scale=.7] (q_0) at (-5.00,1) {$c$};
\node[state, scale=.7] (q_1) [above right of=q_0, xshift=6mm, yshift=2mm ]   {$a^1_c$};
\node[state, scale=.7] (q_3) [below right of=q_1, yshift=2mm] {$a^2_c$};
\node[, scale=.7] (dots) [below right of=q_3, yshift=-3mm] {$\dots$};   
\node[state, scale=.7] (q_2) [below right of=q_0, yshift=-2mm, xshift=3mm] {$a^N_c$};

\draw[<->] (q_0.north) to[out=90, in=180] node [midway, fill=orange!15, scale=.7] {$t^1_c$} (q_1.west);
\draw[<->] (q_0.east) to[out=0, in=180] node [midway, fill=orange!15, scale=.7] {$t^2_c$} (q_3.west);
\draw[<->] (q_0.east) to[out=0, in=180] node [midway, fill=orange!15, scale=.7] {$t^3_c$} (dots.west);
\draw[<->] (q_0.south) to[out=270, in=180]  node [midway,  fill=orange!15, scale=.7] {$t^n_c$} (q_2.west);

% \node[group, fit=(q_0)(q_1)(q_3)(dots)(q_2)] (digraph) {};

% \node[state] (c) at (-5.00,1.20) {$c$};
% \node[state] (a1) at (-3.82,2.02) {$a_c^1$};
% \node[state] (a2) at (-3.35,1.45) {$a_c^2$};
% \node[state] (an) at (-3.82,0.52) {$a_c^{N_c}$};
% \node[font=\scriptsize] (dots) at (-3.02,0.98) {$\cdots$};

% \draw[dbl] (c) to[out=110,in=190] node[midway, above, font=\sffamily\tiny] {$t_c^1$} (a1);
% \draw[dbl] (c) to[out=22,in=188] node[midway, above, font=\sffamily\tiny] {$t_c^2$} (a2);
% \draw[dbl] (c) to[out=-8,in=190] node[midway, below, font=\sffamily\tiny] {$t_c^3$} (dots);
% \draw[dbl] (c) to[out=-105,in=190] node[midway, below, font=\sffamily\tiny] {$t_c^{N_c}$} (an);

\begin{scope}[on background layer]
  \node[draw, dashed, rounded corners=2pt, fill=orange!15,
        fit=(q_0)(q_1)(q_2)(dots), inner sep=7pt] (fsm) {};
\end{scope}

% Intent and speech blocks

\node[peach, minimum width=1.85cm, minimum height=0.84cm, font=\sffamily\small] (intent) at (1,1.42)
  {(5) Intent\\recognition};
\node[box, minimum width=1.55cm, minimum height=0.84cm, font=\sffamily\small] (stt) at (4,1.42){\makecell{Speech-\\to-text}};

% EBuddy components region

\node[box, minimum width=2.45cm, minimum height=0.45cm] (wf1) at (-3.7,-0.8) {Workflow 1};
\node[box, minimum width=2.45cm, minimum height=0.45cm, fill=orange!15] (wf2) at (-3.7,-1.5) {Workflow 2};
\node[] (wfd) at (-3.7,-2) {\dots};
\node[box, minimum width=2.45cm, minimum height=0.45cm] (wfn) at (-3.7,-2.5) {Workflow n}; 
\node[ font=\sffamily\small] (arch) at (-6.3,-2.3){\makecell{(2) FSM\\architecture}};

\begin{scope}[on background layer]
\node[peach, draw, rounded corners=1pt, fit=(wf1)(wfn)(arch), inner sep=5pt] (ebg) {};
\end{scope}

\node[draw, rectangle, fill=orange!15, minimum width=0.96cm, minimum height=2.98cm] (ctrl) at (-1,-1.54) {};
\node[rotate=90, font=\sffamily\small] at (ctrl.center) {(3) Tool controls};

\node[box, fill=orange!15, minimum width=7cm, minimum height=0.50cm] (elabel) at (-4,-3.5) {EBuddy components};

% Tools dashed box
\node[box, minimum width=1.98cm, minimum height=0.44cm] (tool1) at (1.1,-0.38) {Tool 1};
\node[box, minimum width=1.98cm, minimum height=0.44cm] (tool2) at (1.1,-1.03) {Tool 2};
\node[font=\sffamily\small] (tdots) at (1.1,-1.58) {$\cdots$};
\node[box, minimum width=1.98cm, minimum height=0.44cm] (toolm) at (1.1,-2.12) {Tool m};
\node[draw, dashed, rounded corners=1pt, fit=(tool1)(tool2)(tdots)(toolm), inner sep=5pt] (toolbox) {};

% Workspace and UI monitors
\node[box, minimum width=7cm, minimum height=0.50cm] (workspace) at (3.22,-3.5)
  {Workspace (physical + digital)};
\node[peach, minimum width=1.55cm, minimum height=0.84cm, font=\sffamily\small] (ui) at (5.8,-1.8)
  {(4) UI\\monitors};

% Top/context arrows
\draw[<-] (bodyL.north east) to[out=    0,in=180] (role.west);
% \draw[->] (role.east) -- (bodyR.north west);
\draw[->] (role.east) to[out=0,in=90] (headR.north);
% \draw[->] (bodyL.east) -- (fsm.west);
% \draw[->] (bodyL.south east) -- (-5.95,0.05);
% \draw[->] (-5.95,0.05) -- (-5.78,-0.88);

% Speech/intent arrows
\draw[->] (bodyR.west) -- (stt.east) node[midway, font=\sffamily\tiny, fill=white] {\makecell{Voice\\signal}};
\draw[->] (stt.west) -- (intent.east) node[midway, above, font=\sffamily\tiny, text width=1.10cm, align=center]{Transcription};

\draw[->] (intent) -- (ebg) node[midway, right, font=\sffamily\tiny] {Next state};

\draw[->] (stt.west) -- (intent.east) node[midway, above, font=\sffamily\tiny, text width=1.10cm, align=center]{Transcription};
\draw[<-] (intent.west)-- (intent.west-|fsm.east) node[midway, above, font=\sffamily\tiny] {$(c,t_c^1,\ldots,t_c^p)$} node[midway,below, font=\sffamily\tiny] {\makecell{Current ($c$) and\\accessible ($a$)\\states. Available\\transitions ($t$)}};

% Workflow/control/tool arrows
% \draw[->] (fsm.south) -- (-4.45,-0.90) -- (wf1.north);
\draw[<->] (wf1.east) -- (ctrl.west |- wf1);
\draw[<->] (wf2.east) -- (ctrl.west |- wf2);
\draw[<->] (wfn.east) -- (ctrl.west |- wfn);
\draw[<->] (ctrl.east |- tool1.west) -- (tool1.west);
\draw[<->] (ctrl.east |- tool2.west) -- (tool2.west);
\draw[<->] (ctrl.east |- toolm.west) -- (toolm.west);

% Tool/UI/user/workspace arrows
\draw[<-] (tool1.east) -- (bodyR.west);
\draw[<-] (tool2.east) -- (bodyR.west);
\draw[<-] (toolm.east) -- (bodyR.west);
% \draw[->] (toolm.south) -- (toolm.south |- workspace.north);
\draw[<-] (ui.south) -- (ui.south |- workspace.north);
\draw[->] (ui.north) -- (ui.north |- userLabel.south);
\draw[->] (toolbox.south) -- (toolbox.south |- workspace.north);
\draw[<-] (fsm.south) -- (fsm.south |- wf1.north);

\draw[->] (bodyL.east) -- (ebg);
\draw[->] (bodyL.east) -- (fsm);

\end{tikzpicture}
% \end{document}

%% file: diagrams/diagram.tex
\resizebox{\textwidth}{!}{%
\begin{tikzpicture}[
        box/.style={rectangle, draw, rounded corners=2pt, minimum width=3cm, minimum height=1cm, text centered, font=\sffamily, align=center},
        group/.style={draw, dashed, inner sep=5pt, rounded corners=5pt},
        shorten >=1pt,
        node distance=1.5cm,
        auto
        ]
        
    \node[box, fill=cyan!5] (q) {Operator's query $q$};
    
    \node[inner sep=0pt] (topi)    at ($(q)+(-1,.5)$) {};
    \node[inner sep=0pt] (bottomi) at ($(q)+(-1,-4)$) {};
    \node[inner sep=0pt] (topo)    at ($(q)+(18,.5)$) {};
    \node[inner sep=0pt] (bottomo) at ($(q)+(18,-4)$) {};
    
    \node[box, fill=cyan!5] (c) [below of=q] {Current state $c$};
    \node[box, fill=cyan!5] (ts) [below of=c, xshift=5mm] {Available transitions' names\\ to compare to: $t^1_c,t^2_c,\dots,t^n_c$};
    \node[state] (q_0) [right of=c, xshift=15mm]     {$c$};
    \node[state] (q_1) [above right of=q_0, xshift=5mm]   {$a^1_c$};
    \node[state] (q_3) [below right of=q_1, yshift=5mm] {$a^2_c$};
    \node[] (dots) [below right of=q_3] {$\dots$};   
    \node[state] (q_2) [below right of=q_0] {$a^N_c$};

    \node[group, fit=(topi)(bottomi)(c)(dots)] (data sources) {};
    \node at ($(data sources.north west)+(-0.1,0.3)$) [anchor=west] {\small \textit{inputs}};
    
    \node[box, fill=yellow!5] (sim) [right of=data sources, xshift=75mm] {SENTENCE SIMILARITY\\ semantic \& lexical};
    \node[box, fill=green!10] (out1) [right of=q, xshift=167.5mm] {1. transition $t^2_c$, similarity: $74\%$};
    \node[box, fill=green!5] (out2) [below of=out1] {2. transition $t^3_c$, similarity: $32\%$};
    \node[box, fill=green!1] (out3) [below of=out2] {3. transition $t^1_c$, similarity: $12\%$};
    \node[] (out4) [below of=out3, yshift=5mm] {\dots};

    \draw[<->] (q_0.north) to[out=90, in=180] node [midway, fill=white] {$t^1_c$} (q_1.west);
    \draw[<->] (q_0.east) to[out=0, in=180] node [midway, fill=white] {$t^2_c$} (q_3.west);
    \draw[<->] (q_0.east) to[out=0, in=180] node [midway, fill=white] {$t^3_c$} (dots.west);
    \draw[<->] (q_0.south) to[out=270, in=180]  node [midway, fill=white] {$t^n_c$} (q_2.west);

    \node[group, fit=(topo)(bottomo)(out2)] (output) {};
    \node at ($(output.north west)+(-0.1,0.3)$) [anchor=west] {\small \textit{output}};

    \draw[->, thick] (data sources.east) to[out=0,in=180] (sim.west);
    \draw[->, thick] (sim.east) to[out=0,in=180] (output.west);
\end{tikzpicture}
}

%% file: bibliography.bib
@article{bonarini2020communication,
  title={Communication in human-robot interaction},
  author={Bonarini, Andrea},
  journal={Current Robotics Reports},
  volume={1},
  number={4},
  pages={279--285},
  year={2020},
  publisher={Springer}
}

@article{karpus2025human,
  title={Human cooperation with artificial agents varies across countries},
  author={Karpus, Jurgis and Shirai, Risako and Verba, Julia Tovar and Schulte, Rickmer and Weigert, Maximilian and Bahrami, Bahador and Watanabe, Katsumi and Deroy, Ophelia},
  journal={Scientific reports},
  volume={15},
  number={1},
  pages={10000},
  year={2025},
  publisher={Nature Publishing Group UK London}
}

@inproceedings{radford2023robust,
  title={Robust speech recognition via large-scale weak supervision},
  author={Radford, Alec and Kim, Jong Wook and Xu, Tao and Brockman, Greg and McLeavey, Christine and Sutskever, Ilya},
  booktitle={International conference on machine learning},
  pages={28492--28518},
  year={2023},
  organization={PMLR}
}

@article{10.1145/3729236,
author = {Jin, Kebing and Zhuo, Hankz Hankui},
title = {Integrating AI Planning with Natural Language Processing: A Combination of Explicit and Tacit Knowledge},
year = {2025},
issue_date = {August 2025},
publisher = {Association for Computing Machinery},
address = {New York, NY, USA},
volume = {16},
number = {4},
issn = {2157-6904},
url = {https://doi.org/10.1145/3729236},
doi = {10.1145/3729236},
journal = {ACM Trans. Intell. Syst. Technol.},
month = aug,
articleno = {97},
numpages = {37}
}

@misc{web:skillworx,
  author = {{TT PSC}},
  title = {SkillWorx},
  url = {https://ttpsc.com/en/solutions/skillworx/}
}

@misc{web:mcp,
  author = {Anthropic},
  title = {What is the Model Context Protocol (MCP)?},
  url = {https://modelcontextprotocol.io/docs/getting-started/intro}
}

@misc{web:transitions,
  author = {Tal Yarkoni, Alexander Neumann},
  title = {transitions},
  url = {https://github.com/pytransitions/transitions}
}

@INPROCEEDINGS{bussolan2025personalized,
  author={Bussolan, Andrea and Avram, Oliver and Pignata, Andrea and Urgese, Gianvito and Baraldo, Stefano and Valente, Anna},
  booktitle={2025 IEEE International Conference on Engineering, Technology, and Innovation (ICE/ITMC)}, 
  title={Personalized Mental State Evaluation in Human-Robot Interaction using Federated Learning}, 
  year={2025},
  volume={},
  number={},
  pages={1-9},
  doi={10.1109/ICE/ITMC65658.2025.11106641}}

@InProceedings{avram2024advancingcom,
author="Avram, Oliver
and Fasana, Corrado
and Baraldo, Stefano
and Valente, Anna",
editor="Secchi, Cristian
and Marconi, Lorenzo",
title="Advancing Human-Robot Collaboration by Robust Speech Recognition in Smart Manufacturing",
booktitle="European Robotics Forum 2024",
year="2024",
publisher="Springer Nature Switzerland",
address="Cham",
pages="168--173",
isbn="978-3-031-76428-8"
}

@inproceedings{wang2020minilmdeepselfattentiondistillation,
author = {Wang, Wenhui and Wei, Furu and Dong, Li and Bao, Hangbo and Yang, Nan and Zhou, Ming},
title = {MINILM: deep self-attention distillation for task-agnostic compression of pre-trained transformers},
year = {2020},
isbn = {9781713829546},
publisher = {Curran Associates Inc.},
address = {Red Hook, NY, USA},
booktitle = {Proceedings of the 34th International Conference on Neural Information Processing Systems},
articleno = {485},
numpages = {13},
location = {Vancouver, BC, Canada},
series = {NIPS '20}
}

@misc{devlin2019bertpretrainingdeepbidirectional,
      title={BERT: Pre-training of Deep Bidirectional Transformers for Language Understanding}, 
      author={Jacob Devlin and Ming-Wei Chang and Kenton Lee and Kristina Toutanova},
      year={2019},
      eprint={1810.04805},
      archivePrefix={arXiv},
      primaryClass={cs.CL},
      url={https://arxiv.org/abs/1810.04805}, 
}

@article{mehrish2023speechdl,
  author  = {Ambuj Mehrish and Navonil Majumder and Rishabh Bhardwaj and Rada Mihalcea and Soujanya Poria},
  title   = {A Review of Deep Learning Techniques for Speech Processing},
  journal = {Information Fusion},
  year    = {2023},
  doi     = {10.1016/j.inffus.2023.101869},
}

@article{avram2022generalized,
  title={Generalized behavior framework for mobile robots teaming with humans in harsh environments},
  author={Avram, Oliver and Baraldo, Stefano and Valente, Anna},
  journal={Frontiers in Robotics and AI},
  volume={9},
  pages={898366},
  year={2022},
  publisher={Frontiers Media SA}
}

@article{guo2023speech,
  title={Speech-based human-exoskeleton interaction for lower limb motion planning},
  author={Guo, Eddie and Perlette, Christopher and Sharifi, Mojtaba and Grasse, Lukas and Tata, Matthew and Mushahwar, Vivian K and Tavakoli, Mahdi},
  journal={arXiv preprint arXiv:2310.03137},
  year={2023}
}

@misc{leanza2025conceptbotenhancingrobotsautonomy,
      title={ConceptBot: Enhancing Robot's Autonomy through Task Decomposition with Large Language Models and Knowledge Graph}, 
      author={Alessandro Leanza and Angelo Moroncelli and Giuseppe Vizzari and Francesco Braghin and Loris Roveda and Blerina Spahiu},
      year={2025},
      eprint={2509.00570},
      archivePrefix={arXiv},
      primaryClass={cs.RO},
      url={https://arxiv.org/abs/2509.00570}, 
}

@INPROCEEDINGS{DBLP:journals/corr/abs-2109-14718,
  author={Migimatsu, Toki and Bohg, Jeannette},
  booktitle={2022 International Conference on Robotics and Automation (ICRA)}, 
  title={Grounding Predicates through Actions}, 
  year={2022},
  volume={},
  number={},
  pages={3498-3504},
}

@misc{brahman2024plasmamakingsmalllanguage,
      title={PlaSma: Making Small Language Models Better Procedural Knowledge Models for (Counterfactual) Planning}, 
      author={Faeze Brahman and Chandra Bhagavatula and Valentina Pyatkin and Jena D. Hwang and Xiang Lorraine Li and Hirona J. Arai and Soumya Sanyal and Keisuke Sakaguchi and Xiang Ren and Yejin Choi},
      year={2024},
      eprint={2305.19472},
      archivePrefix={arXiv},
      primaryClass={cs.CL},
      url={https://arxiv.org/abs/2305.19472}, 
}

@misc{web:cadai,
  author = {{CADAICO}},
  title = {CADAI Assistant},
  url = {https://www.cadai-platform.co/blog/revolutionizing-cad-design-the-power-of-cadai-assistant}
}

@inproceedings{mallis2025cad,
  title={CAD-assistant: tool-augmented vllms as generic cad task solvers},
  author={Mallis, Dimitrios and Karadeniz, Ahmet Serda and Cavada, Sebastian and Rukhovich, Danila and Foteinopoulou, Niki and Cherenkova, Kseniya and Kacem, Anis and Aouada, Djamila},
  booktitle={Proceedings of the IEEE/CVF International Conference on Computer Vision},
  pages={7284--7294},
  year={2025}
}

@misc{web:claude,
  author = {{Anthropic}},
  title = {Go from prompt to production with Claude Code},
  url = {https://claude.com/product/claude-code}
}

@article{coronado2020,
    author = {Coronado, E. and Mastrogiovanni, F. and Indurkhya, B. and Venture, G.},
    title = {Visual Programming Environments for End-User Development of intelligent and social robots, a systematic review},
    journal = {Journal of Computer Languages},
    year = {2020},
    volume = {58},
    pages = {100970}
}

@article{rodriguez2021,
    author = {Rodriguez-Guerra, D. and Sorrosal, G. and Cabanes, I. and Calleja, C.},
    title = {Human-Robot Interaction Review: Challenges and Solutions for Modern Industrial Environments},
    journal = {Ieee Access},
    year = {2021},
    volume = {9}, pages = {108557-108578}
}

@article{matheson2019,
    author = {Matheson, E. and Minto, R. and Zampieri, E. and Faccio, M. and Rosati, G.}, 
    title = {Human–Robot Collaboration in Manufacturing Applications: A Review}, 
    journal = {Robotics},
    year = {2019},
    volume = {8},
    issue = {4},
    pages = {100}
}
